\documentclass{article}
\usepackage{cite}
\usepackage{amsmath,amssymb,amsfonts}
\usepackage{algorithmic}
\usepackage{graphicx}
\usepackage{textcomp}
\usepackage{url}
\usepackage{xcolor}
\usepackage{fonts-tlwg}
\usepackage{arxiv}
\usepackage[utf8]{inputenc} 
\usepackage[T1]{fontenc}    
\usepackage{hyperref}       
\usepackage{url}            
\usepackage{booktabs}       
\usepackage{amsfonts}       
\usepackage{nicefrac}       
\usepackage{microtype}      
\usepackage{graphicx}

\def\BibTeX{{\rm B\kern-.05em{\sc i\kern-.025em b}\kern-.08em
    T\kern-.1667em\lower.7ex\hbox{E}\kern-.125emX}}
\begin{document}

\title{Thai Wav2Vec2.0 with CommonVoice V8}

\author{
Wannaphong~Phatthiyaphaibun\textsuperscript{\dag}, Chompakorn~Chaksangchaichot\textsuperscript{‡‡}, Peerat Limkonchotiwat\textsuperscript{\dag}, \\
\textbf{Ekapol Chuangsuwanich}\textsuperscript{‡}, \textbf{Sarana Nutanong}\textsuperscript{\dag}\\
  \textsuperscript{\dag}School of Information Science and Technology, VISTEC, Thailand\\
  \textsuperscript{‡‡}VISAI, Thailand \\
  \textsuperscript{‡}Department of Computer Engineering,  Chulalongkorn University, Thailand \\
  \texttt{\{wannaphong.p\_s21,peerat.l\_s19,snutanon\}@vistec.ac.th,}\\
    \texttt{chompakornc\_pro@vistec.ac.th} \\
    \texttt{ekapolc@cp.eng.chula.ac.th}
  }


\maketitle

\begin{abstract}
Recently, Automatic Speech Recognition (ASR), a system that converts audio into text, has caught a lot of attention in the machine learning community. Thus, a lot of publicly available models were released in HuggingFace. However, most of these ASR models are available in English; only a minority of the models are available in Thai. Additionally, most of the Thai ASR models are closed-sourced, and the performance of existing open-sourced models lacks robustness. To address this problem, we train a new ASR model on a pre-trained XLSR-Wav2Vec model with the Thai CommonVoice corpus V8 and train a trigram language model to boost the performance of our ASR model. We hope that our models will be beneficial to individuals and the ASR community in Thailand.\footnote{Our model is publicly available at \url{https://huggingface.co/wannaphong}.}
\end{abstract}


\section{Introduction}
The goal of an Automatic Speech Recognition (ASR) system is to transcribe an audio file to text. It is a domain-specific task, similar to other machine learning work \cite{a1}. However, in order to build a robust ASR system, a large amount of data is required. Because the number of datasets is so limited, low-resource languages like Thai suffer from not having a good ASR system. Several studies were proposed to address this problem by using a self-supervised pre-training method. For example, Wav2Vec2.0 from Facebook \cite{a2} used a lot of unlabeled data to pre-train an unsupervised objective before training an ASR model as a downstream task.

Recently, the Thai ASR community, led by AIResearch.in.th and PyThaiNLP~\cite{a3}, released the Thai Wav2Vec2.0 ASR model by finetuning the XLSR-Wav2Vec2 model with the Thai CommonVoice corpus V7 on the NewMM tokenizer~\cite{a4}. This work states that the CommonVoice dataset has data leakage on their train/test splits as the same speakers were identified in different splits. Thus, they re-split the corpus and trained the model to ensure there was no speaker overlapping among splits. The experimental results showed that the model achieved the word error rate (WER) at 13.643\% on the NewMM tokenization and 8.152\% on the deepcut tokenization~\cite{a9}.
%
%

%
In this technical report, we train a new Thai ASR model by fine-tuning the pre-trained XLSR-Wav2Vec2 model with a newer version of the Commonvoice corpus and changing the stop criteria from character error rate (CER) to WER. In addition, we also train a new language model using a tri-gram language model to improve the performance of the ASR model in the decoder stage.

\section{Methodology}
We used the CommonVoice corpus V8 for training models, but the CommonVoice corpus has data leakage, so we re-split the corpus before training models. We used a pre-trained XLSR-Wav2Vec2 available on Huggingface to fine-tune a model on the CommonVoice corpus V8. The language model was also trained on the same corpus using the train split. Additionally, we also fine-tune the model using different tokenizers to investigate the impact of different open-sourced Thai tokenizers on the ASR model. This section will describe the preprocessing step we applied to the corpus, ASR model setups, and language model creation details.

\label{sec:others}
\subsection{Dataset}
We use the CommonVoice corpus V8 as the main dataset for training and evaluating. The CommonVoice corpus is a crowdsourced speech corpus that collects and validates audio by the crowdsourcing method \cite{a5}. Nevertheless, we noticed that the CommonVoice corpus train/test split has data leakage where the same speakers were identified in both the train and test set. Thus, we re-split the corpus before training the model, following Charin Polpanumas~\cite{a3}.\footnote{The source code and corpus is publicly available at \url{https://github.com/wannaphong/thai_commonvoice_dataset}.}. 
%
We split the CommonVoice V8 dataset as follows:
\begin{enumerate}
\item Removes all data in the CommonVoice corpus V7 from the CommonVoice corpus V8.
\item Splitting the Common Voice corpus V8 without the Common Voice corpus V7.
\item Add the CommonVoice corpus V7 back to the corpus.
\end{enumerate}

With these steps, we can ensure that the same data used in the training and test set of the old CommonVoice corpus V7 split remains the same while the new V8 data is added without speaker leakage between the training and test set. Moreover, we cleaned the corpus according to AIResearch.in.th's work (i.e., removing non-alphanumeric characters). We also fixed missing words and replaced a repeat character (maiyamok) with repeated text by hand.

The statistics of Thai CommonVoice corpus v7 and 8 are shown in Table~\ref{tab:statistic}.
\begin{table}[htbp]
\caption{The statistic of Thai Common Voice v7 and 8}
\centering
\begin{tabular}{|l|l|l|}
\hline
\textbf{Set}   & \textbf{CommonVoice V7}                  & \textbf{CommonVoice V8}                  \\ \hline
Train & 116 hours 18 minutes 41 seconds & 118 hours 45 minutes 35 seconds \\ \hline
Valid & 2 hours 39 minutes 48 seconds   & 5 hours 0 minutes 54 seconds    \\ \hline
Test  & 3 hours 7 minutes 36 seconds    & 5 hours 9 minutes 5 seconds     \\ \hline
Total & 122 hours 6 minutes 5 seconds   & 128 hours 55 minutes 35 seconds \\ \hline
\end{tabular}
\label{tab:statistic}
\end{table}

\subsection{ASR Model}
We fine-tune the \texttt{wav2vec2-large-xlsr-53} model with Connectionist Temporal Classification (CTC)~\cite{a8}. For the model setting, we use the same setting as AIResearch.in.th did, but we train models with two word-tokenizers while NewMM, and Deepcut~\cite{a9}. We evaluate the WER score on the validation set for the model selection to select the best model.



\subsection{Language model}

We train a tri-gram language model with the CommonVoice V8 training set text using KenLM~\cite{a11}. We trained two language models that were tokenized by two differnet tokenizers: NewMM and DeepCut. We then leverage the language model to boost the performance of Wav2Vec2.0, following Patrick von Platen~\cite{a10}.



\section{Experimental Results}

We evaluate the performance of our models using the WER and CER scores~\cite{a12}.
In addition, we perform post-processing similar to the training stage, and retokenize the prediction as follows:
\begin{enumerate}
\item Remove all white space in the predicted text.
\item Retokenize the predicted text using the same word-tokenizers in the training stage.
\item Joined the tokenized words with whitespaces
\item Evaluate the performance of our models using the WER and CER.
\end{enumerate}

\subsection{Thai CommonVoice V8}

The ASR results were shown in Table~\ref{tab:cv8table}. 
Comparing to the baseline models (AIResearch and PyThaiNLP), our models improves the WER from 17.4\% and 11.9\% to 16.3\% on NewMM tokenizer, and 11.4\% on DeepCut.
Moreover, we used the Trigram language model to boost the performance of Wav2Vec2.0, and the WER decreased from 17.4\% to 12.5\%.
We can conclude that the boosting technique from Patrick von Platen~\cite{a10} significantly improves ASR models' performance.

\begin{table}[htbp]
\caption{Thai CommonVoice V8}
\centering
\begin{tabular}{|l|l|l|l|ll}
\cline{1-4}
Model                                  & WER by newmm (\%)  & WER by deepcut (\%) & CER (\%)               &  &  \\ \cline{1-4}
AIResearch.in.th and PyThaiNLP \cite{a3}                                 & 17.414503          & 11.923089           & 3.854153          &  &  \\ \cline{1-4}
wav2vec2 with deepcut                  & 16.354521          & 11.424476           & 3.684060          &  &  \\ \cline{1-4}
wav2vec2 with newmm                    & 16.698299          & 11.436941           & 3.737407          &  &  \\ \cline{1-4}
wav2vec2 with deepcut + language model & 12.630260          & 9.613886            & 3.292073          &  &  \\ \cline{1-4}
wav2vec2 with newmm + language model   & \textbf{12.583706} & \textbf{9.598305}   & \textbf{3.276610} &  &  \\ \cline{1-4}
\end{tabular}
\label{tab:cv8table}
\end{table}

\subsection{Thai CommonVoice v7}
The results is shown on the Table~\ref{tab:cv7table}. The best model is Wav2Vec2.0 with newmm + language model.

\begin{table}[htbp]
\caption{Thai CommonVoice V7}
\begin{tabular}{|l|l|l|l|}
\hline
Model                                  & WER by newmm (\%) & WER by deepcut (\%) & CER (\%)      \\ \hline
AIResearch.in.th and PyThaiNLP \cite{a3}                                 & 13.936698         & 9.347462            & 2.804787 \\ \hline
wav2vec2 with deepcut                  & 12.776381         & 8.773006            & 2.628882 \\ \hline
wav2vec2 with newmm                    & 12.750596         & 8.672616            & 2.623341 \\ \hline
wav2vec2 with deepcut + language model & 9.940050          & 7.423313            & 2.344940 \\ \hline
wav2vec2 with newmm + language model   & \textbf{9.559724} & \textbf{7.339654}   & \textbf{2.277071} \\ \hline
\end{tabular}
\label{tab:cv7table}
\end{table}

\section{Conclusions}
We train Thai Automatic Speech Recognition by fine-tune a pre-trained XLSR-Wav2Vec2 model with a newer CommonVoice corpus while change the early stop criterion from CER to WER. We also train the language model and use it to boost the ASR performance. The best model is wav2vec2 with the NewMM + language model. We achieve a better word error rate than previous works with the help of language model.

\vspace{12pt}
\end{document}